%% file: paper.tex
\documentclass{article}

\usepackage{microtype}
\usepackage{graphicx}
\usepackage{subfigure}
\usepackage{booktabs} % for professional tables
\usepackage{amsmath}
\usepackage{amsfonts}
\usepackage{mathtools}
\usepackage{siunitx}
\usepackage{appendix}

\input{defs.tex}

\newcommand{\rpm}{\raisebox{.2ex}{$\scriptstyle\pm$}}

\usepackage{hyperref}
\usepackage{appendix}

\usepackage[accepted]{icml2018}

\icmltitlerunning{A Note on the Inception Score}

\begin{document}

\twocolumn[
\icmltitle{A Note on the Inception Score}



\icmlsetsymbol{equal}{*}

\begin{icmlauthorlist}
\icmlauthor{Shane Barratt}{equal,to}
\icmlauthor{Rishi Sharma}{equal,to}
\end{icmlauthorlist}

\icmlaffiliation{to}{Stanford University, Stanford, CA}

\icmlcorrespondingauthor{Shane Barratt}{sbarratt@stanford.edu}
\icmlcorrespondingauthor{Rishi Sharma}{rsh@stanford.edu}

\icmlkeywords{Generative adversarial networks, evaluation methods, deep generative models}

\vskip 0.3in
]



\printAffiliationsAndNotice{\icmlEqualContribution} % otherwise use the standard text.

\begin{abstract}
Deep generative models are powerful tools that have produced impressive results in recent years. These advances have been for the most part empirically driven, making it essential that we use high quality evaluation metrics. In this paper, we provide new insights into the Inception Score, a recently proposed and widely used evaluation metric for generative models, and demonstrate that it fails to provide useful guidance when comparing models. We discuss both suboptimalities of the metric itself and issues with its application. Finally, we call for researchers to be more systematic and careful when evaluating and comparing generative models, as the advancement of the field depends upon it.
\end{abstract}

\section{Introduction}
The advent of new deep learning techniques for generative modeling has led to a resurgence of interest in the topic within the artificial intelligence community. Most notably, recent advances have allowed for the generation of hyper-realistic natural images~\citep{karras2017progressive}, in addition to applications in style transfer~\citep{zhu2017cycleGAN, isola2016image2image}, image super-resolution~\citep{ledig2016superresolution}, natural language generation~\citep{guo2017leakGAN}, music generation~\citep{morgen16music}, medical data generation~\citep{esteban2017medical}, and physical modeling~\citep{barati2017transport}. In sum, these applications represent a major advance in the capabilities of machine intelligence and will have significant and immediate practical consequences. Even more promisingly, in the long run, deep generative models are a potential method for developing rich representations of the world from unlabeled data, similar to how humans develop complex mental models, in an unsupervised way, directly from sensory experience. The human ability to imagine and consider potential future scenarios with rich clarity is a crucial feature of our intelligence, and deep generative models may bring us a small step closer to replicating that ability {\it in silico}.

Despite a widespread recognition that high-dimensional generative models lie at the frontier of artificial intelligence research, it remains notoriously difficult to evaluate them. In the absence of meaningful evaluation metrics, it becomes challenging to rigorously make progress towards improved models. As a result, the generative modeling community has developed various ad-hoc evaluative criteria. The Inception Score is one of these ad-hoc metrics that has gained popularity to evalute the quality of generative models for images.

In this paper, we rigorously investigate the most widely used metric for evaluating image-generating models, the Inception Score, and discover several shortcomings within the underlying premise of the score and its application. This metric, while of importance in and of itself, also serves as a paradigm that illustrates many of the difficulties faced when designing an effective method for the evaluation of black-box generative models. In Section~\ref{eval-gen-models}, we briefly review generative models and discuss why evaluating them is often difficult. In Section~\ref{inception-score}, we review the Inception Score and discuss some of its characteristics. In Section~\ref{problems-inception}, we describe what we have identified as the five major shortcomings of the Inception Score, both within the mechanics of the score itself and in the popular usage thereof. We propose some alterations to the metric and its usage to make it more appropriate, but some of the shortcomings are systemic and difficult to eliminate without altering the basic premise of the score.

\section{Evaluating (Black-Box) Generative Models} \label{eval-gen-models}

In generative modeling, we are given a dataset of samples $\mathbf{x}$ drawn from some unknown probability distribution $p_{r}(\mathbf{x})$. The samples $\mathbf{x}$ could be images, text, video, audio, GPS traces, etc. We want to use the samples $\mathbf{\mathbf{x}}$ to derive the unknown real data distribution $p_r(x)$. Our generative model $G$ encodes a distribution over new samples, $p_{g}(\mathbf{x})$. The aim is that we find a generative distribution such that $p_{g}(\mathbf{x}) \approx p_{r}(\mathbf{x})$ according to some metric.

If we are able to directly evaluate $p_g(\mathbf{x})$, then it is common to calculate the likelihood of a held-out dataset under $p_g$ and choose the model that maximizes this likelihood. For most applications, this approach is effective\footnote{It has been shown that log-likelihood evaluation can be misled by simple mixture distributions~\citep{theis2015note, van2015locally}, but this is only relevant in some applications.}. Unfortunately, in many state-of-the-art generative models, we do not have the luxury of an explicit $p_g$. For example, latent variable models like Generative Adversarial Networks (GANs) do not have an explicit representation of the distribution $p_g$, but rather implicitly map random noise vectors to samples through a parameterized neural network~\citep{goodfellow2014generative}. 

Some metrics have been devised that use the structure within an individual class of generative models to compare them~\citep{im2016generative}. However, this makes it impossible to make global comparisons between different classes of generative models. In this paper, we focus on the evaluation of \emph{black-box} generative models where we assume that we can sample from $p_g$ and assume nothing further about the structure of the model.

Many metrics have been proposed for the evaluation of black-box generative models. One way is to approximate a density function over generated samples and then calculate the likelihood of held-out samples. This can be achieved using Parzen Window Estimates as a method for approximating the likelihood when the data consists of images, but other non-parametric density estimation techniques exist for other data types~\citep{breuleux2010parzen}. A more indirect method for evaluation is to apply a pre-trained neural network to generated images and calculate statistics of its output or at a particular hidden layer. This is the approach taken by the Inception Score~\cite{salimans2016improved}, Mode Score~\cite{che2016mode} and Fr\'echet Inception Distance (FID)~\cite{heusel2017gans}. These scores are often motivated by demonstrating that it prefers models that generate realistic and varied images and is correlated with visual quality. Most of the aforementioned metrics can be fooled by algorithms that memorize the training data. Since the Inception Score is the most widely used metric in generative modeling for images, we focus on this metric.

Further, there are several works concerned with the evaluation of evaluation metrics themselves. One study examined several common evaluation metrics and found that the metrics do not correlate with each other. The authors further argue that generative models need to be directly evaluated for the application they are intended for~\cite{theis2015note}. As generative models become integrated into more complex systems, it will be harder to discern their exact application aside from effectively capturing high-dimensional probability distributions thus necessitating high-quality evaluation metrics that are not specific to applications. A recent study investigated several sample-based evaluation metrics and argued that Maximum Mean Discrepancy (MMD) and the 1-Nearest-Neighbour (1-NN) two-sample test satisfied most of the desirable properties of a metric~\cite{empiricalstudy2017}. Further, a recent study found that over several different datasets and metrics, there is no clear evidence to suggest that any model is better than the others, if enough computation is used for hyperparameter search~\cite{lucic2017gans}. This result comes despite the claims of different generative models to demonstrate clear improvements on earlier work (e.g. WGAN as an improvement on the original GAN). In light of the results and discussion in this paper, which casts doubt on the most popular metric used, we do not find the results of this study surprising.

\section{The Inception Score for Image Generation} \label{inception-score}

Suppose we are trying to evaluate a trained generative model $G$ that encodes a distribution $p_g$ over images $\mathbf{\hat{x}}$. We can sample from $p_g$ as many times as we would like, but do not assume that we can directly evaluate $p_g$. The Inception Score is one way to evaluate such a model~\cite{salimans2016improved}. In this section, we re-introduce and motivate the Inception Score as a metric for generative models over images and point out several of its interesting properties.

\subsection{Inception v3}
The Inception v3 Network~\cite{szegedy2016rethinking} is a deep convolutional architecture designed for classification tasks on ImageNet~\cite{deng2009imagenet}, a dataset consisting of 1.2 million RGB images from 1000 classes. Given an image $\mathbf{x}$, the task of the network is to output a class label $y$ in the form of a vector of probabilities $p(y|\mathbf{x}) \in [0,1]^{1000}$, indicating the probability the network assigns to each of the class labels. The Inception v3 network is one of the most widely used networks for transfer learning and pre-trained models are available in most deep learning software libraries.

\subsection{Inception Score} \label{inception score}
The Inception Score is a metric for automatically evaluating the quality of image generative models~\cite{salimans2016improved}. This metric was shown to correlate well with human scoring of the realism of generated images from the CIFAR-10 dataset. The IS uses an Inception v3 Network pre-trained on ImageNet and calculates a statistic of the network's outputs when applied to generated images.

\begin{equation}
\text{IS}(G) = \exp{\big ( \ \Expect_{\mathbf{x} \sim p_g} \  D_{KL} ( \ p(y | \mathbf{x}) \;\|\; p(y) \ ) \ \big )},
\end{equation}

where $\mathbf{x} \sim p_g$ indicates that $\mathbf{x}$ is an image sampled from $p_g$, $D_{KL}(p \| q)$ is the KL-divergence between the distributions $p$ and $q$, $p(y|\mathbf{x})$ is the conditional class distribution, and $p(y) = \int_\mathbf{x} p(y|\mathbf{x})p_g(\mathbf{x})$ is the marginal class distribution. The $\exp$ in the expression is there to make the values easier to compare, so it will be ignored and we will use $\ln(\text{IS}(G))$ without loss of generality.

The authors who proposed the IS aimed to codify two desirable qualities of a generative model into a metric:
\begin{enumerate}
\item The images generated should contain clear objects (i.e. the images are sharp rather than blurry), or $p(y|\mathbf{x})$ should be low entropy. In other words, the Inception Network should be highly confident there is a single object in the image.
\item The generative algorithm should output a high diversity of images from all the different classes in ImageNet, or $p(y)$ should be high entropy.
\end{enumerate}

If both of these traits are satisfied by a generative model, then we expect a large KL-divergence between the distributions $p(y)$ and $p(y|x)$, resulting in a large IS.

\subsection{Digging Deeper into the Inception Score}
Let's see why the proposed score codifies these qualities. The expected KL-divergence between the conditional and marginal distributions of two random variables is equal to their Mutual Information (for proof see Appendix A):

\begin{equation}
\ln(\text{IS}(G))=I(y ; \mathbf{x}).
\label{log_incep_score}
\end{equation}

In other words, the IS can be interpreted as the measure of dependence between the images generated by $G$ and the marginal class distribution over $y$. The Mutual Information of two random variables is further related to their entropies:

\begin{equation}
I(y ; \mathbf{x}) = H(y) - H(y | \mathbf{x} ).
\label{eq:ents}
\end{equation}

This confirms the connection between the IS and our desire for $p(y|\mathbf{x})$ to be low entropy and $p(y)$ to be high entropy. As a consequence of simple properties of entropy we can bound the Inception Score (for proof see Appendix B):

\begin{equation}
1 \leq \text{IS}(G) \leq 1000.
\end{equation}

\subsection{Calculating the Inception Score} \label{calcuate-score}
We can construct an estimator of the Inception Score from samples $\mathbf{x}^{(i)}$ by first constructing an empirical marginal class distribution,

\begin{equation} \label{generator-marginal}
\hat{p}(y) = \frac{1}{N} \sum_{i=1}^{N} p(y | \mathbf{x}^{(i)}),
\end{equation}

where $N$ is the number of sample images taken from the model. Then an approximation to the the expected KL-divergence can be computed by

\begin{equation} \label{estimator}
\text{IS}(G) \approx \exp( \frac{1}{N} \sum_{i=1}^N D_{KL}(p(y | \mathbf{x}^{(i)} \ \| \ \hat{p}(y))).
\end{equation}

The original proposal of the IS recommended applying the above estimator $10$ times with $N=5,000$ and then taking the mean and standard deviation of the resulting scores. At first glance, this procedure seems troubling and in Section~\ref{sec:scorecalculation} we lay out our critique.

\section{Issues With the Inception Score} \label{problems-inception}
As mentioned earlier,~\citet{salimans2016improved} introduced the Inception Score because, in their experiments, it correlated well with human judgment of image quality. Though we don't dispute that this is the case within a significant regime of its usage, there are several problems with the Inception Score that make it an undesirable metric for the evaluation and comparison of generative models. 

Before illustrating in greater detail the problems with the Inception Score, we offer a simple one-dimensional example that illustrates some of its troubles. Suppose our true data comes with equal probability from two classes which have respective normal distributions $N(-1, 2)$ and $N(1, 2)$. The Bayes optimal classifier is $p(y=1|x) = \frac{p(x|y=1)}{p(x|y=0) + p(x|y=1)}$. We can then use this $p(y|x)$ to calculate an analog to the Inception Score in this setting. The optimal generator according to the Inception Score outputs $-\infty$ and $+\infty$ with equal probability, as it achieves $H(y|x)=0$ and $H(y)=\log 2$ and thus an Inception Score of $2$.  Furthermore, many other distributions will also achieve high scores, e.g. the uniform distribution $U(-100, 100)$ and the centered normal distribution $N(0, 20)$, because they will result in $H(y)=\log 2$ and reasonably small $H(y|x)$. However, the true underlying distribution $p(x)$ will achieve a lower score than the aforementioned distributions.

In the general setting, the problems with the Inception Score fall into two categories\footnote{A third issue with the usage of Inception Score is that the code most commonly used to calculate the score has a number of errors, including using an esoteric version of the Inception Network with 1008 classes, rather than the actual 1000. See our GitHub issue for more details:~\url{https://github.com/openai/improved-gan/issues/29}.}:

\begin{enumerate}
    \item Suboptimalities of the Inception Score itself
    \item Problems with the popular usage of the Inception Score
\end{enumerate}

In this section we enumerate both types of issues. In describing the problems with popular usage of the Inception Score, we omit citations so as to not call attention to individual papers for their practices. However, it is not difficult to find many examples of each of the issues we discuss.

\subsection{Suboptimalities of the Inception Score Itself}

\subsubsection{Sensitivity to Weights}
\begin{table*}[t!]
  \caption{Inception Scores on 50k CIFAR-10 training images, 50k ImageNet validation images and ImageNet Validation top-1 accuracy. IV2 TF is the Tensorflow Implementation of the Inception Score using the Inception V2 network. IV3 Torch is the PyTorch implementation of the Inception V3 network~\cite{paszke2017automatic}. IV3 Keras is the Keras implementation of the Inception V3 network~\cite{chollet2015keras}. Scores were calculated using 10 splits of N=5,000 as in the original proposal.}
  \label{inception-scores-weights}
  \centering
  \vskip 0.15in
  \begin{tabular}{llll}
    \toprule
    & \multicolumn{3}{c}{Network}                   \\
    \cmidrule{2-4}
    & IV2 TF & IV3 Torch & IV3 Keras \\
    \midrule
    CIFAR-10                & $11.237 \rpm 0.11$ & $9.737 \rpm 0.148$ & $10.852 \rpm 0.181$ \\
    ImageNet Validation     & $63.028 \rpm 8.311$ & $63.702 \rpm 7.869$ & $65.938 \rpm 8.616$ \\
    Top-1 Accuracy & 0.756 & 0.772 & 0.777 \\
    \bottomrule
  \end{tabular}
\end{table*}

Different training runs of the Inception network on a classification task for ImageNet result in different network weights due to randomness inherent in the training procedure. These differences in network weights typically have minimal effect on the classification accuracy of the network, which speaks to the robustness of the deep convolutional neural network paradigm for classifying images. Although these networks have virtually the same classification accuracy, slight weight changes result in drastically different scores for the exact same set of sampled images. This is illustrated in Table~\ref{inception-scores-weights}, where we calculate the Inception Score for 50k CIFAR-10 training images and 50k ImageNet Validation images using 3 versions of the Inception network, each of which achieve similar ImageNet validation classification accuracies.

The table shows that the mean Inception Score is $3.5\%$ higher for ImageNet validation images, and $11.5\%$ higher for CIFAR validation images, depending on whether a Keras or Torch implementation of the Inception Network are used, both of which have almost identical classification accuracy. The discrepancies are even more pronounced when using the Inception V2 architecture, which is often the network used when calculating the Inception Score in recent papers.

This shows that the Inception Score is sensitive to small changes in network weights that do not affect the final classification accuracy of the network. We would hope that a good metric for evaluating generative models would not be so sensitive to changes that bear no relation to the quality of the images generated. Furthermore, such discrepancies in the Inception Score can easily account for the advances that differentiate ``state-of-the-art'' performance from other work, casting doubt on claims of model superiority.

\subsubsection{Score Calculation and Exponentiation}
\label{sec:scorecalculation}
In Section~\ref{calcuate-score}, we described that the Inception Score is taken by applying the estimator in Equation~\ref{estimator} for $N$ large ($\approx 50,000$). However, the score is not calculated directly for $N = 50,000$, but instead the generated images are broken up into chunks of size $N \over n_\text{splits}$ and the estimator is applied repeatedly on these chunks to compute a mean and standard deviation of the Inception Score. Typically, $n_\text{splits} = 10$. For datasets like ImageNet, where there are 1000 classes in the original dataset, ${N \over n_\text{splits}} = 5000$ samples are not enough to get good statistics on the marginal class distribution of generated images $\hat{p}(y)$ through the method described in Equation~\ref{generator-marginal}.\footnote{ImageNet also has a skew in its class distribution, so we should be careful to train on a subset of ImageNet that has a uniform distribution over classes when applying this metric or account for it in the calculation of the metric.}

Furthermore, by introducing the parameter $n_{\text{splits}}$ we unnecessarily introduce an extra parameter that can change the final score, as shown in Table~\ref{vary-splits}.

\begin{table*}[t!]
  \centering
  \caption{Changing Inception Score as we vary $N$ for Inception v3 in Torch. It is assumed that $50,000$ samples are taken and $N$ represents the size of the splits the Inception Score is averaged over.}
  \label{vary-splits}
  \vskip 0.15in
  \begin{tabular}{lllllllll}
    \toprule
    $n_{\text{splits}}$ & 1 & 2 & 5 & 10 & 20 & 50 & 100 & 200 \\
    \midrule  \\
    mean score & 9.9147 & 9.9091 & 9.8927 & 9.8669 & 9.8144 & 9.6653 & 9.4523 & 9.0884 \\
    standard deviation & 0 & 0.00214 & 0.1010 & 0.1863 & 0.2220 & 0.3075 & 0.3815 & 0.4950 \\
    \bottomrule
  \end{tabular}
\end{table*}

This dependency on $n_{\text{splits}}$ can be removed by computing $\hat{p}(y)$ over the entire generated dataset and by removing the exponential from the calculation of Inception Score, such that the average value will be the same no matter how you choose to batch the generated images. Also, by removing the exponential (which the original authors included only for aesthetic purposes), the Inception Score is now interpretable, in terms of mutual information, as the reduction in uncertainty of an image's ImageNet class given that the image is emitted by the generator $G$.

The new Improved Inception Score is as follows
\begin{equation} 
\label{eq:improved-estimator}
\text{S}(G) = \frac{1}{N} \sum_{i=1}^{N} D_{KL}(p(y | \mathbf{x}^{(i)} \ \| \ \hat{p}(y))
\end{equation}
and it improves both calculation and interpretability of the Inception Score. To calculate the average value, the dataset can be batched into any number of splits without changing the answer, and the variance should be calculated over the entire dataset (i.e. $n_\text{splits} = N$).

\subsection{Problems with Popular Usage of Inception Score}

\subsubsection{Usage beyond ImageNet dataset}
Though this has been pointed out elsewhere~\cite{rosca2017variational}, it is worth restating: applying the Inception Score to generative models trained on datasets other than ImageNet gives misleading results. The most common use of Inception Score on non-ImageNet datsets is for generative models trained on CIFAR-10, because it is quite a bit smaller and more manageable to train on than ImageNet. We have also seen the score used on datasets of bedrooms, flowers, celebrity faces, and more. The original proposal of the Inception Score was for the evaluation of models trained on CIFAR-10.

As discussed in Section~\ref{inception score}, the intuition behind the usefulness of Inception Score lies in its ability to recover good estimates of $p(y)$, the marginal class distribution across the set of generated images $X$, and of $p(y|x)$, the conditional class distribution for generated images $x$. As shown in Table~\ref{cifar-class-distribution}, several of the top 10 predicted classes for CIFAR images are obscure and confusing, suggesting that the predicted marginal distribution $p(y)$ is far from correct and casting doubt on the first assumption underlying the score.

\begin{table}[H]
  \caption{Marginal Class Distribution of Inception v3 on CIFAR vs Actual Class Distribution}
  \label{cifar-class-distribution}
  \centering
  \vskip 0.15in
  \begin{tabular}{ll}
    \toprule
    Top 10 Inception Score Classes & CIFAR-10 Classes \\
    \midrule \\
    Moving Van  &  Airplane \\
    Sorrel (garden herb)  &  Automobile \\
    Container Ship  & Bird \\
    Airliner  & Cat \\
    Threshing Machine  & Deer \\
    Hartebeest (antelope)  & Dog \\
    Amphibian  & Frog \\
    Japanese Spaniel (dog breed)  & Horse \\
    Fox Squirrel  & Ship \\
    Milk Can  &  Truck \\
    \bottomrule
  \end{tabular}
\end{table}

Since the classes in ImageNet and CIFAR-10 do not line up identically, we cannot expect perfect alignment between the classes predicted by the Inception Network and the actual classes within CIFAR-10. Nevertheless, there are many classes in ImageNet that align more appropriately with classes in CIFAR than some of those chosen by the Inception Network. One of the reason for the promotion of bizarre classes (e.g. milk can, fox squirrel) is also that ImageNet contains many more specific categories than CIFAR, and thus the probability of Cat is spread out over the many different breeds of cat, leading to a higher entropy in the conditional distribution. This is another reason that testing on a network trained on a wholly separate dataset is a poor choice.

The second assumption, that the distribution over classes $p(y|x)$ will be low entropy, also does not hold to the degree that we would hope. The average entropy of the conditional distribution $p(y|x)$ conditioned on an image from the training set of CIFAR is 4.664 bits, whereas the average entropy conditioned on a uniformly random image (pixel values uniform between 0 and 255) is 6.512 bits, a modest increase relative to the $\sim10$ bits of entropy possible. For comparison, the average entropy of $p(y|x)$ conditioned on images in the ImageNet validation set is 1.97 bits. As such, the entropy of the conditional class distribution on CIFAR is closer to that of random images than to the actual images in ImageNet, casting doubt on the second assumption underlying the Inception Score.

Given the premise of the score, it makes quite a bit more sense to use the Inception Score only when the Inception Network has been trained on the same dataset as the generative model. Thus the original Inception Score should be used only for ImageNet generators, and its variants should use models trained on the specific dataset in question.

\subsubsection{Optimizing the Inception Score (indirectly \& implicitly)}
As mentioned in the original proposal, the Inception Score should only be used as a ``rough guide'' to evaluating generative models, and directly optimizing the score will lead to the generation of adversarial examples~\cite{szegedy2013intriguing}. It should also be noted that optimizing the metric indirectly by using it for model selection will similarly tend to produce models that, though they may achieve a higher Inception Score, tend toward adversarial examples. % more than they tend towards expressive generative models <--- that's a bold claim
It is not uncommon in the literature to see algorithms use the Inception Score as a metric to optimize early stopping, hyperparameter tuning, or even model architecture. Furthermore, by promoting models that achieve high Inception Scores, the generative modeling community similarly optimizes implicitly towards adversarial examples, though this effect will likely only be significant if the Inception Score continues to be optimized for within the community over a long time scale.

\begin{figure}[t]
  \centering
  \includegraphics[scale=.68]{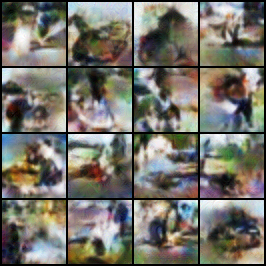} %0.325
  \caption{Sample of generated images achieving an Inception Score of $900.15$. The maximum achievable Inception Score is 1000, and the highest achieved in the literature is on the order of 10.}
  \label{fig:realistic}
\end{figure}

In Appendix~\ref{sec:high-inception} we show how to achieve high inception scores by gently altering the output of a WGAN to create examples that achieve a nearly perfect Inception Score, despite looking no more like natural images than the original WGAN output. A few such images are shown in Figure~\ref{fig:realistic}, which achieve an Inception Score of 900.15.

\subsubsection{Not Reporting Overfitting}
It is clear that a generative algorithm that memorized an appropriate subset of the training data would perform extremely well in terms of Inception Score, and in some sense we can treat the score of a validation set as an upper bound on the possible performance of a generative algorithm. Thus, it is extremely important when reporting the Inception Score of an algorithm to include some alternative score demonstrating that the model is not overfitting to training data, validating that the high score achieved is not simply replaying the training data. Nevertheless, in many works the Inception Score is treated as a holistic metric that can summarize the performance of the algorithm in a single number. In the generative modeling community, we should not use the existence of a metric that correlates with human judgment as an excuse to exclude more thorough analysis of the generative technique in question.

\section{Conclusion}
Deep learning is an empirical subject. 
In an empirical subject, success is determined by using evaluation metrics--developed and accepted by researchers within the community--to measure performance on tasks that capture the essential difficulty of the problem at hand. Thus, it is crucial to have meaningful evaluation metrics in order to make scientific progress in deep learning.
An outstanding example of successful empirical research within machine learning is the Large Scale Visual Recognition Challenge benchmark for computer vision tasks that has arguably produced most of the greatest computer vision advances of the last decade\cite{russakovsky2015imagenet}.
This competition has and continues to serve as a perfect sandbox to develop, test, and verify hypotheses about visual recognition systems. 
Developing common tasks and evaluative criteria can be more difficult outside such narrow domains as visual recognition, but we think it is worthwhile for generative modeling researchers to devote more time to rigorous and consistent evaluative methodologies. 
This paper marks an attempt to better understand popular evaluative methodologies and make the evaluation of generative models more consistent and thorough.

In this note, we highlighted a number of suboptimalities of the Inception Score and explicated some of the difficulties in designing a good metric for evaluating generative models. Given that our metrics to evaluate generative models are far from perfect, it is important that generative modeling researchers continue to devote significant energy to the evaluation and validations of new techniques and methods.

\section*{Acknowledgements}

This material is based upon work supported by the National Science Foundation Graduate Research Fellowship under Grant No. DGE-1656518.

\bibliography{paper}
\bibliographystyle{icml2018}

\begin{appendix}

\section*{Proof of Equation~\ref{log_incep_score}}

\begin{equation}
\ln(\text{Inception Score}(G)) = \Expect_{\mathbf{x} \sim p_g} [D_{KL} ( \ p(y | \mathbf{x}) \;\|\; p(y) \ ]
\label{proof_a_1}
\end{equation}

\begin{equation}
= \sum_{\mathbf{x}} p_g(\mathbf{x}) D_{KL} ( \ p(y | \mathbf{x}) \;\|\; p(y) \ )
\label{proof_a_2}
\end{equation}

\begin{equation}
= \sum_{\mathbf{x}} p_g(\mathbf{x}) \sum_i p(y=i | \mathbf{x}) \ln \frac{ p(y=i | \mathbf{x})}{ p(y=i)}
\label{proof_a_3}
\end{equation}

\begin{equation}
= \sum_{\mathbf{x}} \sum_i p(\mathbf{x}, y=i) \ln \frac{p(\mathbf{x}, y=i)}{p(\mathbf{x})p(y=i)}
\label{proof_a_4}
\end{equation}

\begin{equation}
= I(y; \mathbf{x})
\label{proof_a_5}
\end{equation}

where Equation \ref{proof_a_1} is the definition of the Inception Score, Equation \ref{proof_a_2} expands the expectation, Equation \ref{proof_a_3} uses the definition of the KL-divergence, Equation \ref{proof_a_4} uses the definition of conditional probability twice and Equation \ref{proof_a_5} uses the definition of the Mutual Information.

\section*{Proof of Equation~\ref{eq:ents}}
We can derive an upper bound of Equation~\ref{eq:ents},

\begin{equation}
H(y) - H(y | \mathbf{x}) \leq H(y) \leq \ln(1000).
\label{incepscoremax}
\end{equation}

The first inequality is because entropy is always positive and the second inequality is because the highest entropy discrete distribution is the uniform distribution, which has entropy $\ln(1000)$ as there are 1000 classes in ImageNet. Taking the exponential of our upper bound on the log IS, we find that the maximum possible IS is {\bf 1000}. We can also find a lower bound

\begin{equation}
H(y) - H(y | \mathbf{x}) \geq 0,
\end{equation}

because the conditional entropy $H(y | \mathbf{x})$ is always less than the unconditional entropy $H(y)$. Again, taking the exponential of our lower bound, we find that the minimum possible IS is {\bf 1}. We can combine our two inequalities to get the final expression,

\begin{equation}
1 \leq \text{IS}(G) \leq 1000.
\end{equation}

\section*{Achieving High Inception Scores} \label{sec:high-inception}

\begin{algorithm}[t!]
   \caption{Optimize Generator.}
   \label{alg:gen}
\begin{algorithmic}[1]
   \STATE {\bfseries Require:} $\epsilon$, the learning rate. $P(\mathbf{x})$, a distribution over initial images. $N$, the number of iterations to run the inner-optimization procedure. $j$, the last class outputted by the generator.
   \STATE Sample $\mathbf{x}$ from $P(\mathbf{x})$.
   \REPEAT
    \STATE $\mathbf{x} \leftarrow \mathbf{x} +  \epsilon \cdot \text{sgn} (\nabla_{\mathbf{x}} p(y=j | \mathbf{x}))$
   \UNTIL{$\mathbf{x}$ converged}
   \STATE $j \leftarrow (j + 1)\mod 1000$ 
   \STATE return $\mathbf{x}$
\end{algorithmic}
\end{algorithm}

We repeat Equation \ref{incepscoremax} here for the convenience of the reader

$$ \ln(\text{Inception Score}(G))=H(y) - H(y | \mathbf{x}) \leq \ln(1000) $$

It should be relatively clear now how we can achieve an Inception score of $1000$. We require the following:

\begin{enumerate}
\item $H(y)=\ln(1000)$. We can achieve this by making $p(y)$ the uniform distribution.
\item $H(y|\mathbf{x})=0$. We can achieve this by making $p(y=i|\mathbf{x})$=1 for one $i$ and $0$ for all of the others.
\end{enumerate}

Since the Inception Network is differentiable, we have access to the gradient of the output with respect to the input $\nabla_{\mathbf{x}} p(y=i|\mathbf{x})$. We can then use this gradient to repeatedly update our image to force $p(y=i | \mathbf{x})=1$. 

Let's make this more concrete. Given a class $i$, we can sample an image $\mathbf{x}$ from some distribution $P(\mathbf{x})$, then repeatedly update $\mathbf{x}$ to maximize $p(y=i | \mathbf{x})$ for some $i$. Our resulting generator cycles from $i=1$ to $1000$ repeatedly, outputting the image that is the result of the above optimization procedure. This procedure is identical to the Fast Gradient Sign Method (FGSM) for adversarial attacks against neural networks\cite{goodfellow2014explaining}. In the original proposal of the Inception Score, the authors noted that directly optimizing it would lead to adversarial examples\cite{salimans2016improved}.

In theory, it should achieve a near perfect Inception Score as long as $N\epsilon$ is suitably large enough. The full generative algorithm is summarized in Algorithm~\ref{alg:gen}. We note that the replay attack is equivalent to $P(\mathbf{x})$ being the empirical distribution of the training data and $N$ or $\epsilon$ being equal to $0$. 

\begin{figure}
\centering
\subfigure[]{\label{fig:a}\includegraphics[width=.4\linewidth]{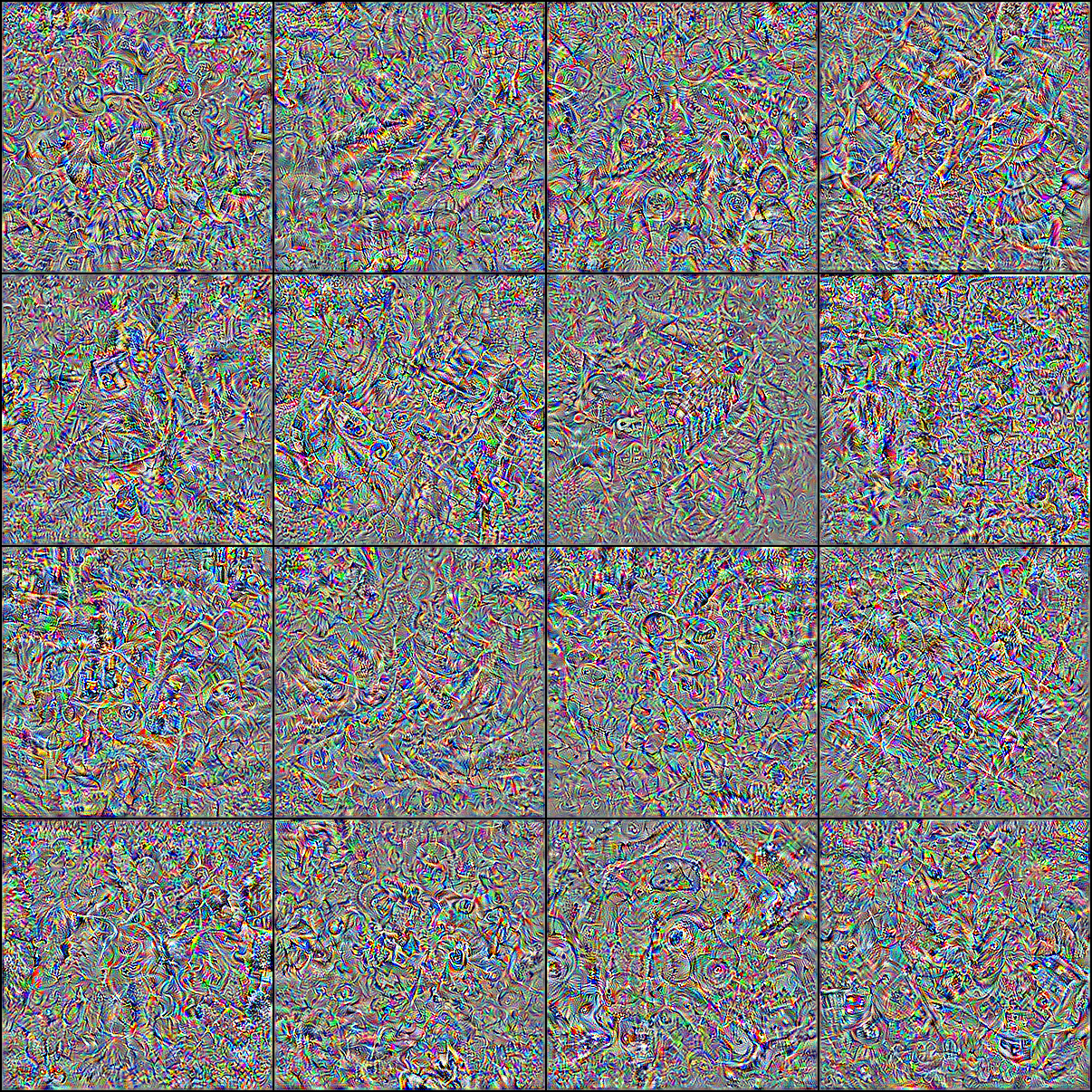}}
\subfigure[]{\label{fig:b}\includegraphics[width=.4\linewidth]{figs/test-1.png}}
\caption{Sample images from generative algorithms that achieve nearly optimal Inception Scores. (a), sample images from random initializations with gradient fine-tuning. (b), sample images from WGAN initializations with gradient fine-tuning.}
\label{fig:generators}
\end{figure}

We can realize this algorithm by setting $\epsilon=.001$, $N=100$ and $P(\mathbf{x})$ to be a uniform distribution over images. The resulting generator achieves produces images shown in the left of Figure~\ref{fig:generators} and an Inception score of $986.10$.

We can make the images more realistic by making $P(\mathbf{x})$ a pre-trained Wasserstein GAN (WGAN)~\citep{arjovsky2017wasserstein} trained on CIFAR-10. This method produces realistic-looking examples that achieve a near-perfect Inception Score, shown in the right of Figure~\ref{fig:generators} and an Inception score of $900.10$.

\end{appendix}
\end{document}

%% file: defs.tex
\newcommand{\Expect}{\mathbb{E}}